\documentclass{article}

\usepackage[nonatbib,preprint]{neurips_2020}

\usepackage[utf8]{inputenc} 
\usepackage[T1]{fontenc}    
\usepackage{hyperref}       
\usepackage{url}            
\usepackage{booktabs}       
\usepackage{amsfonts}       
\usepackage{nicefrac}       
\usepackage{microtype}      
\usepackage{amsmath}
\usepackage{amsfonts}
\usepackage{amsmath}
\DeclareMathOperator*{\argmin}{arg\,min}
\usepackage{mathtools}
\usepackage{esdiff}
\usepackage{cleveref}
\usepackage{amsmath,bm,mathtools,upgreek}
\usepackage{amssymb}
\usepackage{amsfonts}
\usepackage{amsmath}
\usepackage{makecell}
\usepackage{mathtools}
\usepackage{mathrsfs}
\usepackage{scalerel}
\usepackage{algorithm}
\usepackage[noend]{algpseudocode}
\usepackage{amsmath,bm,mathtools,upgreek}
\usepackage{amssymb}
\usepackage{amsfonts}
\usepackage{amsmath}
\usepackage{makecell}
\usepackage{mathtools}
\usepackage{mathrsfs}
\usepackage{scalerel}
\usepackage{algorithm}
\usepackage{makecell}
\usepackage{mathtools}
\usepackage{caption}

\usepackage{graphicx}   

\title{Regularized L21-Based Semi-NonNegative Matrix Factorization }

\author{Anthony D. Rhodes\\Portland State University\\\texttt{arhodes@pdx.edu} \And Bin Jiang\\Portland State University\\ \texttt{bjiang@pdx.edu}}

\usepackage[T1]{fontenc}
\usepackage[
backend=biber
]
{biblatex}
\begin{document}

\maketitle

\begin{abstract}
We present a general-purpose data compression algorithm, Regularized L21 Semi-NonNegative Matrix Factorization (L21 SNF). L21 SNF provides robust, parts-based compression applicable to mixed-sign data for which high fidelity, individual data point reconstruction is paramount. We derive a rigorous proof of convergence of our algorithm. Through experiments, we show the use-case advantages presented by L21 SNF, including application to the compression of highly overdetermined systems encountered broadly across many general machine learning processes.
   
\end{abstract}

\section{Background}
\noindent Data reduction algorithms represent an essential component of machine learning systems. The use of such reductions are supported by topological properties of data, including the well-known Manifold Hypothesis \cite{NIPS2010_3958}.

\noindent Our work presents a novel compression algorithm which renders a parts-based compression applicable to mixed-sign data for which high fidelity, individual data point reconstruction is paramount. We achieve this result by solving a constrained optimization problem for matrix reconstruction with respect to L2-1 loss. We present the details of our algorithm in Section 2. In Section 3 we demonstrate proof of convergence, and in Section 4 we provide experimental results.

\noindent As a precursor to our work, we consider the Non-negative Matrix Factorization (NMF) framework \cite{lee99, NIPS2000_1861, 4359171, NIPS2005_2757, 10.1016/j.csda.2008.01.011, Kong}. NMF is defined as the problem of finding a matrix factorization of a given non-negative matrix $\mathbf{X}^{m \times n}$ so that $\mathbf{X}\approx \mathbf{WH}$ for non-negative factors $\mathbf{W}^{m \times r}$ and $\mathbf{H}^{r \times n}$; compression is achieved when $r < min(m,n)$.

\noindent A significant feature of this methodology is that it gives rise to a parts-based decomposition of $\mathbf{X}.$ In this way, each column of $\mathbf{W}$ represents a basis element in the reduced space; the columns of $\mathbf{H}$ can correspondingly be interpreted as embodying the coordinates for each basis element that render an approximation of the columns of $\mathbf{X}$. Since the number of basis vectors ($r$) is often relatively small, this set of vectors represents a useful latent structure in the data. Finally, because each component in the factorization is restricted to be non-negative, their interaction in approximating $\mathbf{X}$ is strictly \textit{additive}, meaning that the columns of $\mathbf{W}$ yield a parts-based, compressed decomposition of $\mathbf{X}$.

\noindent Lee and Seung \cite{lee99, NIPS2000_1861} provided a well-known solution to NMF in the form of two multiplicative
algorithms based on the standard square Euclidean distance and "divergence", respectively,
defined:
\begin{equation}
\|\mathbf{X}-\mathbf{WH}\|_{F}^{2}=\sum_{ij}(\mathbf{X}_{ij}-\mathbf{WH}_{ij})^{2}
\label{l2norm}
\end{equation}

\begin{equation}
D(\mathbf{X}||\mathbf{WH}) = \sum_{ij} (\mathbf{X}_{ij}\text{log} \frac{\mathbf{X}_{ij}}{\mathbf{WH}_{ij}}-\mathbf{X}_{ij}+\mathbf{WH}_{ij})
\label{divnorm}
\end{equation}
where both the distance and divergence are bounded below by zero, and vanish when $\mathbf{X=WH}$. Divergence furthermore reduces to KL-Divergence \cite{kullback1951} when $\sum_{ij}\mathbf{X}_{ij}=\sum_{ij}\mathbf{WH}_{ij}=1$. Subsequent to \cite{NIPS2000_1861}, related research for NMF has used projected gradients \cite{10.1162/neco.2007.19.10.2756}, non-negative least square \cite{10.1007/s10618-012-0265-y}, and neural approaches \cite{Vu2016CombiningNM}, among other methods.

\section{L21 SNF} 
\noindent When the data matrix $\mathbf{X}$ is not strictly non-negative, NMF is inapplicable. Nevertheless, in many common use cases, a parts-based decomposition is a desideratum for data compression with mixed-sign data. \cite{4685898} introduce a useful compromise toward this end, the Frobenius norm based Semi-Nonnegative Matrix Factorization, in which one (and only one) of the factor matrices is constrained to be non-negative. However, the Frobenius norm in \eqref{l2norm} is known for its instability with respect to noise and outliers \cite{Kong, Liu}.

\noindent In place of the loss functions given by \eqref{l2norm} and \eqref{divnorm}, we propose to instead employ a generally more robust measure that leverages together L2 and L1 loss, termed L2-1 loss \cite{NIPS2010_3988}. The definition of the L2-1 norm is as follows: 
\begin{equation}
\|\mathbf{X} \|_{2,1}=\sum_{i=1}^{n} \sqrt{ \sum_{j=1}^{m}\mathbf{X}_{ji}^{2} }
\label{l21norm}
\end{equation}
\noindent We accordingly define L2-1 loss for matrix factorization by: 
\begin{equation}
\|\mathbf{X-WH} \|_{2,1}=\sum_{i=1}^{n} \sqrt{ \sum_{j=1}^{m}\mathbf{(X}_{ji}-\mathbf({\mathbf{WH}})_{ji})^{2} }
\label{l21normerror}
\end{equation}
\noindent Note in particular that we define L2-1 loss as a sum of L2 vector magnitudes with respect to each column of $\mathbf{X}$. When applied, for example, to a set of convolutional filters \cite{denton} (consider each column of $\mathbf{X}$ as a "flattened" filter), L2-1 loss can viewed as a robust measure that weighs the distance per filter component using L2 cost, while summing over filters with L1 cost. \cite{Kong} use the L2-1 norm to solve NMF, but their method cannot be directly extended to accommodate mixed-sign data.

\noindent Let $\mathbf{X} \in \mathbb{R}^{m\times n},\mathbf{W} \in \mathbb{R}^{m\times k},\mathbf{H} \in \mathbb{R}_+^{k\times n}$; we define the optimization problem framing L2-1 semi non-negative matrix factorization: \\
\begin{equation}
\argmin_{\mathbf{W},\mathbf{H}} \|\mathbf{X}-\mathbf{WH}\|_{2,1}+\overline{\alpha}\|\mathbf{W} \|_{2}^{2}\text{     subject to }  \mathbf{H}\geq 0
\label{optimization}
\end{equation} \\
where $\overline{\alpha}=\frac{\alpha}{2}$ (for simplicity of notation) and $\alpha\geq 0$ is a hyperparameter.

\noindent We present the following Regularized L21 SNF algorithm which provides an iterative solution to \eqref{optimization}; in the following section we prove convergence of our algorithm. 

\renewcommand{\thealgorithm}{}
\begin{algorithm}[H]
\begin{algorithmic}
\State Initialize $\mathbf{H}(0)$ as non-negative matrix, initialize $\mathbf{W}(0)$ (e.g. use k-means)
\For {$t$ in $0:T-1$} \\
{
{\text{   } (1) $\mathbf{H}_{ij}(t+1)=\mathbf{H}_{ij}(t)\sqrt{\frac{(\Phi^{+}\mathbf{D}(t))_{ij}+(\Omega^{-}\mathbf{H}(t)\mathbf{D}(t))_{ij}}
{(\Phi^{-}\mathbf{D}(t))_{ij}+(\Omega^{+}\mathbf{H}(t)\mathbf{D}(t))_{ij}}}$
}\\
 \text{   } $(2)\mathbf{W}(t+1)=[\mathbf{XD}(t)\mathbf{H}(t)^{T}][\alpha\mathbf{I}+\mathbf{H}(t)\mathbf{D}(t)\mathbf{H}(t)^{T}]^{-1}$ } 
 \EndFor
where $\mathbf{W}^{T}(t)\mathbf{W}(t)=\Omega,$ $\Omega=\Omega^{+}-\Omega^{-}$, 
$\mathbf{W}^{T}(t)\mathbf{X}=\Phi,$  $\Phi=\Phi^{+}-\Phi^{-}$, and $\mathbf{D}(t)_{ii}=1/ \|\mathbf{x}^{(i)}-\mathbf{W}(t)\mathbf{h}(t)^{(i)} \|_{2}; \mathbf{x}^{(i)} \in \mathbb{R}^{m\times 1}$ denotes the \textit{i}th column of $\mathbf{X}$, and $\mathbf{h}^{(i)} \in \mathbb{R}_+^{k\times 1}$ denotes the \textit{i}th column of $ \mathbf{H}$. $\Omega^{+}$, $\Omega^{-}$ are positive and negative entries of $\Omega$.
 \caption{Regularized L21 SNF}
 \end{algorithmic}
\end{algorithm}

\section{Proof of Convergence}
\noindent We define the \textit{proxy loss} function using matrix trace: \\
\begin{gather}
\begin{split}
\mathscr{L}(\mathbf{X},\mathbf{WH})=tr[(\mathbf{X}-\mathbf{WH})\mathbf{D}(\mathbf{X}-\mathbf{WH})^{T}]+\alpha tr[\mathbf{W}^{T}\mathbf{W}]
\\ \text{where } \mathbf{D} \in \mathbb{R}^{n\times n},
\mathbf{D}_{ii}= 1/\|\mathbf{x}^{(i)}-\mathbf{W}\mathbf{h}^{(i)} \|_{2} 
\end{split}
\label{proxyloss}
\end{gather}

\noindent Next, we subsequently derive iterative update formulas based on the loss function given in \eqref{proxyloss} and show that these updates incur a monotonic loss in \eqref{optimization}.\\
\begin{equation}
\mathscr{L}(\mathbf{X},\mathbf{WH})=
tr[\mathbf{XDX}^{T}]-2tr[\mathbf{W}^{T}\mathbf{XD}\mathbf{H}^{T}]+tr[\mathbf{WHDH}^{T}\mathbf{W}^{T}]+\alpha tr[\mathbf{W}^{T}\mathbf{W}]
\label{proxyloss2}
\end{equation} \\
Solving $\nabla_{\mathbf{W}}\mathscr{L}=0$, gives the solution:
\begin{equation}
\mathbf{W}=[\mathbf{XDH}^{T}][\alpha\mathbf{I}+\mathbf{HD}\mathbf{H}^{T}]^{-1}
\label{wformula}
\end{equation}

\noindent Now we prove that optimality of \eqref{wformula} by demonstrating that \eqref{proxyloss} is convex; we first consider $\frac{\partial\mathscr{L}}{\partial W_{ij}}$:
\begin{equation}
\frac{\partial\mathscr{L}}{\partial W_{ij}}=2(\mathbf{WHD}\mathbf{H}^{T})_{ij}-2(\mathbf{XDH}^{T})_{ij}+2\alpha(\mathbf{W})_{ij}
\label{lwij}
\end{equation} \\
\noindent Expanding the first term on the RHS of \eqref{lwij} renders the following simplification. 
\begin{equation}
\frac{\partial\mathscr{L}}{\partial W_{ij}}=2\sum_{l=1}^{k}\mathbf{W}_{il}(\mathbf{HD}\mathbf{H}^{T})_{lj}-2(\mathbf{XDH}^{T})_{ij}+2\alpha(\mathbf{W})_{ij}
\label{lwij2}
\end{equation} \\
\noindent The Hessian of $\mathscr{L}$ is consequently: 
\begin{equation}
\frac{\partial\mathscr{L}}{\partial W_{ij}\partial W_{pq}}
=2(\mathbf{HD}\mathbf{H}^{T}+\alpha\mathbf{I})_{qj}\delta_{ip} \quad \quad 1\leq i,p \leq m \quad 1\leq j,q \leq k
\label{lwijpq}
\end{equation}

\noindent Therefore, the Hessian of $\mathscr{L}$ is a block diagonal matrix with each block being of the form $2\mathbf{HDH}^T+2\alpha \mathbf{I}$. It follows that $\mathscr{L}$ is convex, thus the formula given for $\mathbf{W}$ in \eqref{wformula} minimizes $\mathscr{L}$ in \eqref{proxyloss}.

\noindent The previous derivation of \eqref{wformula} and the associated demonstration of optimality furnish a proof for the following Lemma. We now consider \eqref{wformula} as an iterative update rule at step $t$, where we regard $\mathbf{H}(t)$ as fixed at the time of the $t$-th update for $\mathbf{W}$, denoted by $\mathbf{W}(t+1)$. Define $\mathbf{D}(t)_{ii}= 1/\|\mathbf{x}^{(i)}-\mathbf{W}(t)\mathbf{h}(t)^{(i)} \|_{2}$, which depends on $\mathbf{W}(t)$ and is also regarded as fixed at the time of $t$-th update for $\mathbf{W}(t+1)$. The iterative update for matrix $\mathbf{W}$ is given by: 
\begin{equation}
\mathbf{W}(t+1)=[\mathbf{XD}(t)\mathbf{H}(t)^{T}][\alpha \mathbf{I}+\mathbf{H}(t)\mathbf{D}(t)\mathbf{H}(t)^{T}]^{-1}
\label{wtplus1}
\end{equation}

\noindent \textbf{Lemma 1.} Let $\mathbf{W}(t)$ and $\mathbf{W}(t+1)$ represent consecutive updates for $\mathbf{W}$ as prescribed by \eqref{wtplus1}. Under this updating rule, the following inequality holds: 
\begin{gather}
\begin{split}
tr[(\mathbf{X}-\mathbf{W}(t+1)\mathbf{H}(t))\mathbf{D}(t)(\mathbf{X}-\mathbf{W}(t+1)\mathbf{H}(t))^{T}] +\alpha tr[\mathbf{W}^{T}(t+1)\mathbf{W}(t+1)]\\
\leq tr[(\mathbf{X}-\mathbf{W}(t)\mathbf{H}(t))\mathbf{D}(t)(\mathbf{X}-\mathbf{W}(t)\mathbf{H}(t))^{T})]+\alpha tr[\mathbf{W}^{T}(t)\mathbf{W}(t)]\\
\end{split}
\label{lemma1}
\end{gather}\\
\textbf{Proof.} The proof of Lemma 1 follows directly from the optimality of the update formula in \eqref{wtplus1}. $\square$ \\

\noindent \textbf{Lemma 2.} Under the update rule of \eqref{wtplus1}, the following inequality holds where $\overline{\alpha}=\frac{\alpha}{2}$:
\begin{gather}
\begin{split}
\left(\|\mathbf{X}-\mathbf{W}(t+1)\mathbf{H}(t)\|_{2,1}-\|\mathbf{X}-\mathbf{W}(t)\mathbf{H}(t)\|_{2,1}\right) \\
+\overline{\alpha} tr[\mathbf{W}(t+1)\mathbf{W}^T(t+1)]-\overline{\alpha} tr[\mathbf{W}(t)\mathbf{W}^T(t)]\\
\leq \frac{1}{2}\big\{ tr[(\mathbf{X}-\mathbf{W}(t+1)\mathbf{H}(t))\mathbf{D}(t)(\mathbf{X}-\mathbf{W}(t+1)(\mathbf{H}(t))^T] \\ -tr[(\mathbf{X}-\mathbf{W}(t)\mathbf{H}(t))\mathbf{D}(t)(\mathbf{X}-\mathbf{W}(t)\mathbf{H}(t))^T] \big\} \\
+\overline{\alpha} tr[\mathbf{W}(t+1)\mathbf{W}^T(t+1)]-\overline{\alpha} tr[\mathbf{W}(t)\mathbf{W}^T(t)] 
\end{split}
\label{lemma2}
\end{gather}
\textbf{Proof.} The proof is analogous to the proof of Lemma 3 in \cite{Kong}, so we skip it here for brevity.  $\square$ \\

\noindent \textbf{Theorem 1.} Updating $\mathbf{W}$ using formula \eqref{wtplus1} while fixing $\mathbf{H}$ yields a monotonic decrease in the objective function defined by \eqref{optimization}. \\ \\
\textbf{Proof.} By Lemma 1, the right hand side expression in Lemma 2 satisfies:
\begin{gather}
\begin{split}
 \frac{1}{2}tr[(\mathbf{X}-\mathbf{W}(t+1)\mathbf{H}(t))\mathbf{D}(t)(\mathbf{X}-\mathbf{W}(t+1)(\mathbf{H}(t))^T]+\overline{\alpha} tr[\mathbf{W}(t+1)\mathbf{W}^T(t+1)] \\
 - \frac{1}{2}tr[(\mathbf{X}-\mathbf{W}(t)\mathbf{H}(t))\mathbf{D}(t)(\mathbf{X}-\mathbf{W}(t)\mathbf{H}(t))^T]
 -\overline{\alpha} tr[\mathbf{W}(t)\mathbf{W}^T(t)] \leq 0. 
\end{split}
\label{theo1rhs}
\end{gather}
So does the left hand side expression in Lemma 2:
\begin{gather}
\begin{split}
 \|\mathbf{X}-\mathbf{W}(t+1)\mathbf{H}(t)\|_{2,1} + \overline{\alpha} tr[\mathbf{W}(t+1)\mathbf{W}^T(t+1)] \\
 -\|\mathbf{X}-\mathbf{W}(t)\mathbf{H}(t)\|_{2,1}
 -\overline{\alpha} tr[\mathbf{W}(t)\mathbf{W}^T(t)] \leq 0.
\end{split}
\label{theo1lhs}
\end{gather}
Thus proving Theorem 1. $\square$\\

\noindent Next we derive an iterative update formula for $\mathbf{H}$, with $\mathbf{H} \geq 0$; subsequently we prove convergence of this update rule by showing that the proxy loss $\mathscr{L}(\mathbf{X},\mathbf{WH})$ given in \eqref{proxyloss} is monotonically decreasing for fixed $\mathbf{W}$. Since the second term of \eqref{proxyloss}, $\alpha tr[\mathbf{W}^{T}\mathbf{W}]$, is fixed during the $\mathbf{H}$ update, we ignore it here. Define the corresponding \textit{truncated proxy loss}: $F(\mathbf{H})=tr[(\mathbf{X}-\mathbf{WH})\mathbf{D}(\mathbf{X}-\mathbf{WH})^{T}]$. 

\noindent Based on formula \eqref{proxyloss2} as well as $tr(\mathbf{AB})=tr(\mathbf{BA})$, $F(\mathbf{H})$ can be further simplified to
\begin{equation}
F(\mathbf{H})=
tr[\mathbf{XDX}^{T}]-2tr[\mathbf{H}^{T}\mathbf{W}^{T}\mathbf{XD}]+tr[\mathbf{W}^{T}\mathbf{WHDH}^{T}]
\label{fhformula}
\end{equation}

\noindent To prove convergence for $\mathbf{H}$, we utilize an auxiliary function, denoted $\mathscr{A}(\mathbf{H},\mathbf{H'})$ as in \cite{NIPS2000_1861,10.1016/j.csda.2008.01.011}. \\

\textbf{Definition.}  $\mathscr{A}$  is an auxiliary function for $F(\mathbf{H})$ if:
\begin{equation}
\mathscr{A}(\mathbf{H},\mathbf{H'})\geq F(\mathbf{H}), \quad \mathscr{A}(\mathbf{H},\mathbf{H})= F(\mathbf{H})
\label{auxiliarydef}
\end{equation}
\textbf{Lemma 3.}  If $\mathscr{A}$  is an auxiliary function of $F(\mathbf{H})$, then $F(\mathbf{H})$ is non-increasing under the update:
\begin{equation}
\mathbf{H^{t+1}}=\argmin_{\mathbf{H}} \mathscr{A}(\mathbf{H}, \mathbf{H^{t}})
\label{lemma3}
\end{equation}
\textbf{Proof.} 
\begin{equation}
F(\mathbf{H^{t+1}})\leq \mathscr{A}(\mathbf{H^{t+1}},\mathbf{H^{t}})\leq \mathscr{A}(\mathbf{H^{t}},\mathbf{H^{t}})=F(\mathbf{H^{t}}).
\label{auxiliaryproof}
\end{equation}

\noindent We now consider an explicit solution for $\mathbf{H}$ in the form of an iterative update, for which we subsequently prove convergence. Since $\mathbf{H}$ is non-negative, it is helpful to decompose both the $k \times k$ matrix $\mathbf{W}^{T}\mathbf{W}=\Omega$ and the $k \times n$ matrix  $\mathbf{W}^{T}\mathbf{X}=\Phi$ into their positive and negative entries as follows:
\begin{equation}
\Omega^{+}_{ij}=\frac{1}{2}(|\Omega_{ij}|+\Omega_{ij}), \quad
 \Omega^{-}_{ij}=\frac{1}{2}(|\Omega_{ij}|-\Omega_{ij}).
 \label{omegapositive}
\end{equation}
\noindent \textbf{Lemma 4.} Under the iterative update:
\begin{equation}
\mathbf{H}_{ij}(t+1)=\mathbf{H}_{ij}(t)\sqrt{\frac{(\Phi^{+}\mathbf{D}(t))_{ij}+(\Omega^{-}\mathbf{H}(t)\mathbf{D}(t))_{ij}}
{(\Phi^{-}\mathbf{D}(t))_{ij}+(\Omega^{+}\mathbf{H}(t)\mathbf{D}(t))_{ij}}}
\label{lemma4}
\end{equation}
where $\mathbf{W}^{T}(t)\mathbf{W}(t)=\Omega,$ $\Omega=\Omega^{+}-\Omega^{-}$, 
$\mathbf{W}^{T}(t)\mathbf{X}=\Phi$,  $\Phi=\Phi^{+}-\Phi^{-}$, and $\mathbf{D}(t)_{ii}= 1/\|\mathbf{x}^{(i)}-\mathbf{W}(t)\mathbf{h}(t)^{(i)} \|_{2} $, 
the following relation holds for some auxiliary function $\mathscr{A}(\mathbf{H},\mathbf{H}')$:
\begin{equation}
\mathbf{H}(t+1)=\argmin_{\mathbf{H}} \mathscr{A}(\mathbf{H},\mathbf{H}(t))
\label{htplus1}
\end{equation}
\textbf{Proof.} Using the notation introduced above, the truncated proxy loss $F(\mathbf{H})$ in \eqref{fhformula} can be rewritten in the following form: 
\begin{gather}
\begin{split}
F(\mathbf{H})=tr[\mathbf{XD}\mathbf{X}^{T}]
-2tr[\mathbf{H}^{T}\mathbf{\Phi}^{+}\mathbf{D}]+2tr[\mathbf{H}^{T}\mathbf{\Phi}^{-}\mathbf{D}]\\
+tr[\Omega^{+}\mathbf{H}\mathbf{D}\mathbf{H}^{T}]-tr[\Omega^{-}\mathbf{H}\mathbf{D}\mathbf{H}^{T}]
\end{split}
\label{fhterms}
\end{gather}

\noindent In the subsequent steps we provide an auxiliary function $\mathscr{A}(\mathbf{H},\mathbf{H}')$ for $F(\mathbf{H})$. Following \cite{NIPS2010_3988}, in order to construct an auxiliary function that furnishes an upper-bound for $F(\mathbf{H})$, we define $\mathscr{A}(\mathbf{H},\mathbf{H})$ as a sum comprised of terms that represent upper-bounds for each of the positive terms appearing in \eqref{fhterms} and lower-bounds for each of the negative terms, respectively. \\

First, we derive a lower-bound for the second term of \eqref{fhterms}, using $a \geq 1+log\text{ }a, \text{  } \forall \text{  } a>0$:
\begin{equation}
tr[\mathbf{H}^{T}\mathbf{\Phi}^{+}\mathbf{D}]=\sum_{ij}\mathbf{H}_{ij} (\mathbf{\Phi}^{+}\mathbf{D})_{ij} \geq\sum_{ij}(\mathbf{\Phi}^{+}\mathbf{D})_{ij}\mathbf{H}'_{ij}(1+log\frac{\mathbf{H}_{ij}}{\mathbf{H}'_{ij}})
\label{fh2ndterm}
\end{equation}

Second, using the fact that $a\leq \frac{a^2+b^2}{2b} \text{ } \forall$ $a,b>0$, we derive an upper bound for the third term on the RHS of \eqref{fhterms}: 
\begin{equation}
tr[\mathbf{H}^{T}\mathbf{\Phi}^{-}\mathbf{D}]=\sum_{ij}\mathbf{H}_{ij} (\mathbf{\Phi}^{-} \mathbf{D})_{ij}\leq
 \sum_{ij}(\mathbf{\Phi}^{-} \mathbf{D})_{ij}
 \frac{(\mathbf{H}_{ij})^2+(\mathbf{H}'_{ij})^2}{2\mathbf{H'}_{ij}}
 \label{fh3rdterm}
\end{equation}

Third, we apply the following inequality \cite{4685898} to bound the fourth term on the RHS of \eqref{fhterms}.\\

\noindent \textbf{Proposition 1.} For any matrices $\mathbf{A}\in \mathbb{R}^{n \times n}_{+}$, $\mathbf{B}\in \mathbb{R}^{k \times k}_{+}$, 
$\mathbf{S}\in \mathbb{R}^{n \times k}_{+}$, $\mathbf{S}'\in \mathbb{R}^{n \times k}_{+}$, with $\mathbf{A}$ and $\mathbf{B}$ symmetric:
\begin{equation}
tr[\mathbf{S}^{T}\mathbf{ASB}] \leq \sum_{i=1}^{n} \sum_{p=1}^{k} \frac{{(\mathbf{AS'B})}_{ip}\mathbf{S}^{2}_{ip}}{\mathbf{S}'_{ip}}
\label{proposition1}
\end{equation}

Considering the fourth term of the RHS of \eqref{fhterms}, we have: 
\begin{equation}
tr[\Omega^{+}\mathbf{HD}\mathbf{H}^{T}]=tr[\mathbf{H}^{T}\Omega^{+}\mathbf{HD}]
\leq \sum_{ij}\frac{(\Omega^{+}\mathbf{H}'\mathbf{D})_{ij}\mathbf{H}^2_{ij}}{\mathbf{H}'_{ij}}
\label{fh4thterm}
\end{equation}
Finally, we consider the fifth term on the RHS of equation \eqref{fhterms}.

\noindent \textbf{Proposition 2.}
\begin{equation}
tr[\Omega^{-}\mathbf{HD}\mathbf{H}^{T}]\geq \sum_{ijk}\Omega^{-}_{ik}\mathbf{H}'_{kj}\mathbf{D}_{jj}\mathbf{H}'_{ij}\Big(1+log\frac{\mathbf{H}_{kj}\mathbf{H}_{ij}}{\mathbf{H'}_{kj}\mathbf{H}'_{ij}} \Big)
\label{proposition2}
\end{equation}
\textbf{Proof.} 
\begin{gather}
\begin{split}
tr[\Omega^{-}\mathbf{HD}\mathbf{H}^{T}]=tr[\mathbf{H}^{T}\Omega^{-}\mathbf{HD}]=\sum_{ij}(\Omega^{-}\mathbf{HD})_{ij}\mathbf{H}_{ij} \\
= \sum_{ijk}\Omega^{-}_{ik}(\mathbf{HD})_{kj}\mathbf{H}_{ij} = \sum_{ijk}\Omega^{-}_{ik}\mathbf{H}_{kj}\mathbf{D}_{jj}\mathbf{H}_{ij}
\end{split}
\label{proposition2proof1}
\end{gather}
Once again we employ the inequality $a \geq 1+log\text{ }a$, whereupon:
\begin{equation}
tr[\Omega^{-}\mathbf{HD}\mathbf{H}^{T}]\geq \sum_{ijk}\Omega^{-}_{ik}\mathbf{H}'_{kj}\mathbf{D}_{jj}\mathbf{H}'_{ij}\Big(1+log\frac{\mathbf{H}_{kj}\mathbf{H}_{ij}}{\mathbf{H'}_{kj}\mathbf{H}'_{ij}} \Big)
\label{proposition2proof2}
\end{equation}
Thus proving Proposition 2. $\square$

\noindent Putting all the formulas \eqref{fh2ndterm}, \eqref{fh3rdterm}, \eqref{fh4thterm} and \eqref{proposition2} together, we define the auxiliary function $\mathscr{A}(\mathbf{H},\mathbf{H}')$:
\begin{gather}
\begin{split}
\mathscr{A}(\mathbf{H},\mathbf{H}')=tr[\mathbf{XD}\mathbf{X}^{T}]
+2\sum_{ij}(\mathbf{\Phi}^{-}\mathbf{D})_{ij}\frac{(\mathbf{H}_{ij})^2+(\mathbf{H}'_{ij})^2}{2\mathbf{H'}_{ij}}
+\sum_{ij}\frac{(\Omega^{+}\mathbf{H}'\mathbf{D})_{ij}\mathbf{H}^2_{ij}}{\mathbf{H}'_{ij}}\\
 -2\sum_{ij}(\mathbf{\Phi}^{+}\mathbf{D})_{ij}\mathbf{H}'_{ij}(1+log\frac{\mathbf{H}_{ij}}{\mathbf{H}'_{ij}})
-\sum_{ijk}\Omega^{-}_{ik}\mathbf{H}'_{kj}\mathbf{D}_{jj}\mathbf{H}'_{ij}\Big(1+log\frac{\mathbf{H}_{kj}\mathbf{H}_{ij}}{\mathbf{H'}_{kj}\mathbf{H}'_{ij}} \Big)
\end{split}
\label{lhhprime}
\end{gather}
\noindent Observe that $\mathscr{A}(\mathbf{H},\mathbf{H'})\geq F(\mathbf{H})$ and $\mathscr{A}(\mathbf{H},\mathbf{H})= F(\mathbf{H})$, as required
for an auxiliary function, where $F(\mathbf{H})$ denotes the truncated proxy loss as defined in equation \eqref{fhterms}. By the aforementioned Lemma, it follows
that $F(\mathbf{H})$ is non-increasing under the update: $\mathbf{H}(t+1)=\argmin_{\mathbf{H}} \mathscr{A}(\mathbf{H},\mathbf{H}(t))$.

\noindent We now demonstrate that the minimum of $\mathscr{A}(\mathbf{H},\mathbf{H}')$ coincides with the update rule in \eqref{lemma4}. Since
\begin{gather}
\begin{split}
    \frac{\partial\mathscr{A}(\mathbf{H},\mathbf{H}')}{\partial\mathbf{H}_{ij}}   =2(\mathbf{\Phi^{-}D})_{ij}\Big(\frac{\mathbf{H}_{ij}}{\mathbf{H}'_{ij}}\Big)
    +2\frac{(\Omega^{+}\mathbf{H'D})_{ij}\mathbf{H}_{ij}}{\mathbf{H'}_{ij}} \\
   -2(\mathbf{\Phi}^{+}\mathbf{D})_{ij}\Big(\frac{\mathbf{H'}_{ij}}{\mathbf{H}_{ij}}\Big)
    -2\frac{(\Omega^{-}\mathbf{H'D})_{ij}\mathbf{H}'_{ij}}{\mathbf{H}_{ij}} =0\\
\end{split}
\label{lhhprimedev}
\end{gather}
\noindent We solve for $\mathbf{H}_{ij}$, arriving at the update formula given in \eqref{lemma4}. Thus \eqref{lemma4} corresponds with a critical point for $\mathscr{A}(\mathbf{H},\mathbf{H'})$.

\noindent Computing the corresponding Hessian of  $\mathscr{A}(\mathbf{H},\mathbf{H'})$ yields: 
  \begin{equation}
    \frac{\partial\mathscr{A}(\mathbf{H},\mathbf{H'})}{\partial \mathbf{H}_{ij}\partial \mathbf{H}_{kl}}=
    \begin{cases} 
     2\frac{(\Phi^{-}\mathbf{D})_{ij}}{\mathbf{H'}_{ij}}
    +2\frac{(\Omega^{+}\mathbf{H'D})_{ij}}{\mathbf{H}'_{ij}}
    +2\frac{(\Phi^{+}\mathbf{D})_{ij}\mathbf{H'}_{ij}}{\mathbf{H}_{ij}^2}
    +2\frac{(\Omega^{-}\mathbf{H'D})_{ij}\mathbf{H'_{ij}}}{\mathbf{H}_{ij}^2} \quad \text{ if } (i,j)==(k,l) \\
     0 \hspace{8.6cm} \text{otherwise}
    \end{cases}
    \label{lhhprimehessian}
\end{equation}

Hence $\mathscr{A}(\mathbf{H},\mathbf{H'})$ is convex, as was to be shown.
 
\noindent Finally, to conclude the proof of Lemma 4, we show that the iterative update formula given by \eqref{lemma4} additionally enforces non-negativity for the matrix $\mathbf{H}$. To this end, we define a matrix $\mathbf{\Lambda} \in \mathbb{R}^{k \times n}$ of Lagrangian multipliers. This gives the following associated Lagrangian: 
 \begin{gather}
\begin{split}
F(\mathbf{H})_{\Lambda}=tr[\mathbf{XD}\mathbf{X}^{T}]
-2tr[\mathbf{H}^{T}\mathbf{\Phi}^{+}\mathbf{D}]+2tr[\mathbf{H}^{T}\mathbf{\Phi}^{-}\mathbf{D}]\\
+tr[\Omega^{+}\mathbf{H}\mathbf{D}\mathbf{H}^{T}]-tr[\Omega^{-}\mathbf{H}\mathbf{D}\mathbf{H}^{T}]- \Lambda \odot \mathbf{H}
\end{split}
\label{lagrangian}
\end{gather}
where $\odot$ denotes the Hadamard product. The gradient of the Lagrangian is therefore: \\
\begin{equation}
\nabla_{H}F(\mathbf{H})_{\Lambda}=-2\Phi^{+}\mathbf{D}+2\Phi^{-}\mathbf{D}+2\Omega^{+}\mathbf{HD}-2\Omega^{-}\mathbf{HD}-\Lambda
\label{lagrangiangrad}
\end{equation}

\noindent The Karush-Kuhn-Tucker (KKT) conditions \cite{10.5555/1355334} dictate that a necessary condition for optimality with the prescribed non-negative constraints is $\mathbf{H}^{*}\odot \mathbf{\Lambda}=0$, where $\mathbf{H}^{*}$ is optimal. This indicates that an optimal solution necessarily satisfies: 
\begin{equation}
-\Phi^{+}\mathbf{D}+\Phi^{-}\mathbf{D}+\Omega^{+}\mathbf{HD}-\Omega^{-}\mathbf{HD}-\frac{1}{2}\mathbf{\Lambda}=0
\label{kkt1}
\end{equation}
which implies the following by the KKT slackness condition: 
\begin{equation}
\mathbf{H}_{ij}(-\Phi^{+}\mathbf{D}+\Phi^{-}\mathbf{D}+\Omega^{+}\mathbf{HD}-\Omega^{-}\mathbf{HD})_{ij}=0
\label{kkt2}
\end{equation}
Equivalently, the optimal solution satisfies:
\begin{equation}
\mathbf{H}_{ij}^2(-\Phi^{+}\mathbf{D}+\Phi^{-}\mathbf{D}+\Omega^{+}\mathbf{HD}-\Omega^{-}\mathbf{HD})_{ij}=0
\label{kktslackness}
\end{equation}
Solving \eqref{kktslackness} for $\mathbf{H}_{ij}$ renders formula \eqref{lemma4}. This concludes the proof of Lemma 4. $\square$ \\

\noindent \textbf{Lemma 5.} Let $H(t)$ and $H(t+1)$ represent consecutive updates for $\mathbf{H}$ as prescribed by \eqref{lemma4}. Under this updating rule, the following inequality holds: 
\begin{gather}
\begin{split}
tr[(\mathbf{X}-\mathbf{W}(t)\mathbf{H}(t+1))\mathbf{D}(t)(\mathbf{X}-\mathbf{W}(t)\mathbf{H}(t+1))^{T}] \\
\leq tr[(\mathbf{X}-\mathbf{W}(t)\mathbf{H}(t))\mathbf{D}(t)(\mathbf{X}-\mathbf{W}(t)\mathbf{H}(t))^{T})]
\end{split}
\label{lemma5}
\end{gather}
\textbf{Proof.} The proof of Lemma 5 follows directly from Lemma 4 and Lemma 3. $\square$ \\

\noindent \textbf{Lemma 6.} Under the update rule of \eqref{lemma4}, the following inequality holds: 
\begin{gather}
\begin{split}
\|\mathbf{X}-\mathbf{W}(t)\mathbf{H}(t+1)\|_{2,1}-\|\mathbf{X}-\mathbf{W}(t)\mathbf{H}(t)\|_{2,1} \\
\leq \frac{1}{2}\Big[ tr[(\mathbf{X}-\mathbf{W}(t)\mathbf{H}(t+1))\mathbf{D}(t)(\mathbf{X}-\mathbf{W}(t)(\mathbf{H}(t+1))^T] - \\ tr[(\mathbf{X}-\mathbf{W}(t)\mathbf{H}(t))\mathbf{D}(t)(\mathbf{X}-\mathbf{W}(t)\mathbf{H}(t))^T] \Big]
\end{split}
\label{lemma6}
\end{gather}

\textbf{Proof.} The proof of Lemma 6 follows analogously from the proof of Lemma 2. 

\noindent \textbf{Theorem 2.} Updating $\mathbf{H}$ using formula\eqref{lemma4} while fixing $\mathbf{W}$ yields a monotonic decrease in the objective function defined by (5). \\ \\
\noindent \textbf{Proof.} The proof of Theorem 2 is similar to Theorem 1, via Lemma 5 and Lemma 6. 

\section{Experimental Results}
\noindent We perform two general experiments to compare the performance of L21 SNF Algorithm with SNF \cite{4685898}: (1) general data compression via matrix factorization, and (2) qualitative facial image data reconstruction via matrix factorization. To compare general data compression performance, we begin with randomized, mixed-sign data matrices $\mathbf{X}$ (in the range $[-20,20]$) of dimension $10,000 \times 128$. In each case, we perform different degrees of compression; through separate trials, we reduce $\mathbf{X}$ to dimension ${10,000 \times 64}$, ${10,000 \times 32}$, ${10,000 \times 16}$, and ${10,000 \times 8}$. In particular, these extreme matrix dimensions are inspired by the potential use-case applications of highly overdetermined systems (e.g., deep CNN compression, genomic data compression, etc.). \\

\noindent Using cluster-based initialization schemes for $\mathbf{W}$ and $\mathbf{H}$, we compare reconstruction loss using (2) different metrics: (i) normalized Frobenius loss (NFL), i.e., $\frac{\|X-WH\|_F}{\|X\|_F}$ and (ii) normalized L21 loss (NL21) , i.e. $\frac{\|X-WH\|_{2,1}}{\|X\|_{2,1}}$. For initialization, we run K-means for five iterations; this yields initial cluster centroids and indicators. We initialize the basis matrix $\mathbf{W}$ to the rendered cluster centroids; we set the initial coordinates instantiated by $\mathbf{H}$ per the basis set to 1.2 when the datum belongs to the cluster, and 0.2 otherwise. Table 1 and Figure 1 summarizes these findings. Across our experiments we optimize the regularization hyperparameter $\alpha $ using random search \cite{journals/jmlr/BergstraB12} over the interval [0,1]. \\

\noindent Overall, the L21 SNF algorithm demonstrates a substantial improvement in comparison with SNF \cite{4685898} and PCA in reducing L21-based reconstruction loss across each of our experiments, while at the same time maintaining generally strong results for L2-based reconstruction loss (see Figure 1, Table 1). In particular, L21 SNF exhibits significant gains in the case of severely overdetermined systems. In experimental trials of reducing random, mixed-sign matrices of initial dimension $10,000 \times 128$, for instance, L21 SNF shows a relative improvement of 26\% over SNF for the 50\% compression task, while exhibiting only a 4\% increase in L2 loss, comparatively; similarly, for the 75\% compression task, L21 SNF demonstrates an 11\% relative improvement over SNF with respect to L21 loss, and only a 1\% increase in L2 loss compared with SNF. \\

\noindent Lastly, we compare compression quality rendered by L21 SNF with SNF and PCA for the task of compression on a batch of images. For this experiment, we randomly sampled 200 images from the \textit{Large-scale CelebFaces Attributes} (CelebA) dataset \cite{10.1109/ICCV.2015.425}. Each image is of dimension ${89 \times 108}$; we flattened and concatenated this batch of images, rendering a data matrix $\mathbf{X}$ of dimension ${9,612 \times 200}$. We then ran each of the L21 SNF, SNF, and PCA algorithms for 250 iterations, reducing the original matrix to size  ${9,612 \times 100}$. The results of this experiment are shown in Figure 2. \\

\noindent Figure 2 in particular provides a qualitative illustration of the stark contrast in performance among L21 SNF, SNF \cite{4685898} and PCA for compression applied to severely overdetermined datasets. While the reconstruction fidelity for L21 SNF is comparable with the original images from the CelebA dataset, both the SNF and PCA techniques performed poorly by comparison, as each introduces a significant amount of distortion and image artifacts in the reconstruction process.

\section{Discussion}
\noindent We present a new, robust data compression algorithm which renders a parts-based compression of mixed-sign data. Theorems 1 and 2 furnish proofs of the convergence of the iterative updates given by our L21 SNF algorithm. Through experiments, we demonstrate the use-case advantages of our algorithm over the classic NMF and SNF algorithms, particularly in the case of highly overdetermined systems. In future work, we aim to generalize these results to more complex constraints, including sparsity. We anticipate that our algorithm can potentially be applied to a variety of relevant real-world applications in the future, including the compression of deep CNNs, problems in computational biology, and general clustering paradigms.

\begin{table}
\caption{Experimental Results}
\begin{tabular}{ |c|c|c|c|c|c| } 
\hline
Compression & NFL (ours) &  NL21 (ours) & NFL (SNF) & NL21 (SNF) \\
\hline
${10k \times 64}$ & 0.704 (0.694) & $\mathbf{0.498 \text{ } (0.498)}$& 0.674 (0.673) &  $\mathbf{0.672 \text{ }(0.620)}$ \\ 
${10k \times 32}$ & 0.865 (0.855)   & $\mathbf{0.749 \text{ } (0.749) }$ & 0.845 (0.846) &   $\mathbf{0.845 \text{ }(0.844)}$ \\ ${10k \times 16}$ &   0.935 (0.929)   &  0.874 (0.874)& 0.925 (0.924) &    0.924 (0.923)  \\ ${10k \times 8}$ &  0.968 (0.964)  &  0.937 (0.937) & 0.962 (0.962) &  0.962 (0.962)\\ 
\hline
\end{tabular} 
\caption*{Summary of loss measures for L21 SNF algorithm (ours) vs SNF run for 100 iterations, beginning with random, mixed-sign matrix of dimension ${10,000 \times 128}$.}
\end{table}

\begin{figure}
    \centering
    \includegraphics[width=0.70\textwidth]{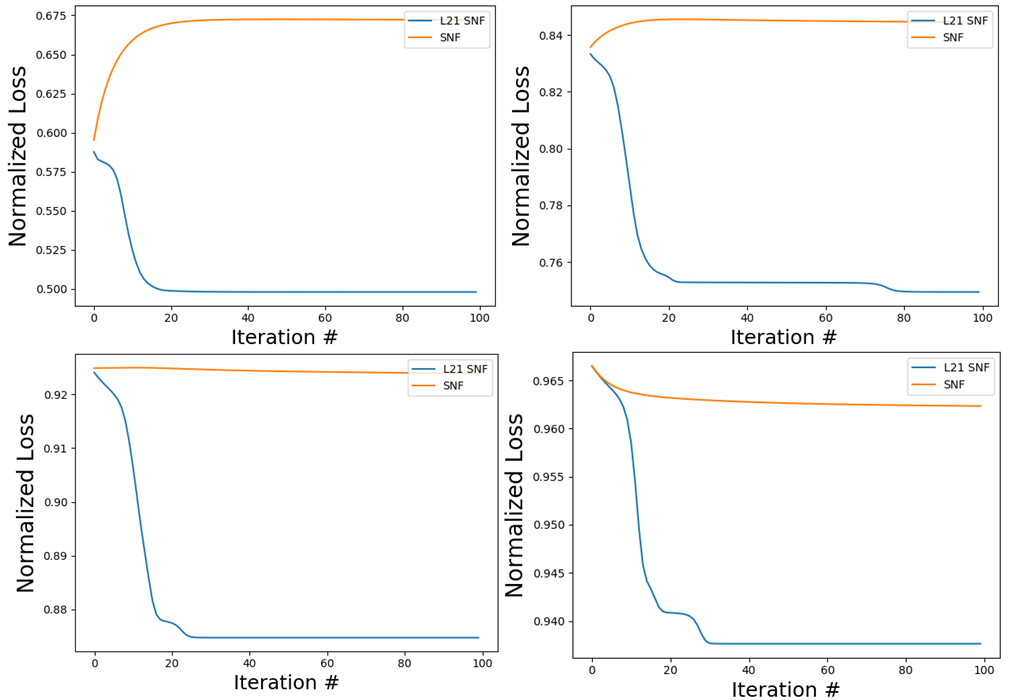} 
    \caption{Comparison of normalized L21 loss for L21 SNF (ours) vs SNF algorithm for compression of matrix $\mathbf{X}$ of dimension ${10,000 \times 128}$: (i) Top-Left, ${10,000 \times 64}$ compression, (ii) Top-Right, ${10,000 \times 32}$, (iii) Bottom-Left, ${10,000 \times 16}$, and (iv) Bottom-Right, ${10,000 \times 8}.$ }
\end{figure}

\begin{figure}
    \centering
    \includegraphics[width=0.75\textwidth]{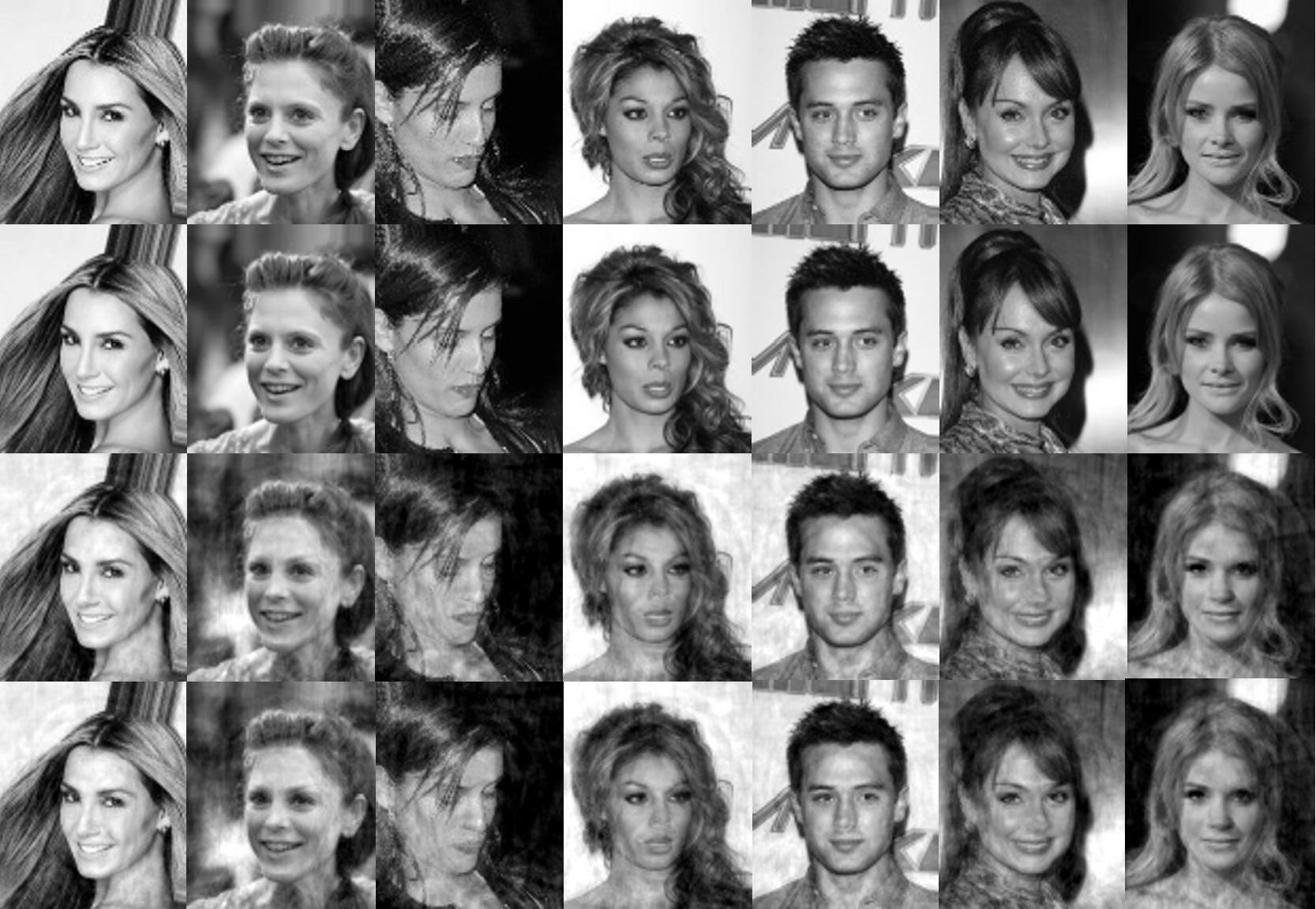} 
    \caption{Results for compression of batch of 200 face images sampled from the CelebA \cite{10.1109/ICCV.2015.425} dataset; we show a sample of seven randomly selected images after compression. The original image batch of dimension ${9,612 \times 200}$ was compressed to ${9,612 \times 100}$; each algorithm was run for 250 iterations. Top: ground-truth images; Second from Top: L21 SNF (ours) rendered result; Second from Bottom: SNF results; Bottom: PCA results. }
\end{figure}

\printbibliography
\end{document}